\documentclass[10pt,journal,compsoc]{IEEEtran}

\usepackage{cite}
\usepackage{graphicx}
\usepackage{amsmath,amssymb,amsfonts}
\usepackage{multirow}
\usepackage{enumitem}
\usepackage{subcaption}
\usepackage{float}
\usepackage{algorithm}
\usepackage{algpseudocode}
\usepackage{color}
\usepackage{booktabs}
\usepackage{lineno}
\usepackage{url}

\begin{document}
\title{Spatio-Temporal Dual Graph Neural Networks for Travel Time Estimation}
\author{Guangyin Jin*,
        Huan Yan*,
        Fuxian Li,
        Jincai Huang,
        and Yong~Li,~\IEEEmembership{Senior Member,~IEEE}
\IEEEcompsocitemizethanks{\IEEEcompsocthanksitem H.~Yan, F.~Li, Y.~Li are with Beijing National Research Center for Information Science and Technology (BNRist), Department of Electronic Engineering, Tsinghua University, Beijing 100084, China.\protect\\
E-mail: lifx19@mails.tsinghua.edu.cn, fengj12ee@hotmail.com, yanhuanthu@gmail.com, \{jindp, liyong07\}@tsinghua.edu.cn
\protect\\
\IEEEcompsocthanksitem G.~Jin and J.~Huang are with College of Systems Engineering, National University of Defense Technology, Changsha, China.\protect\\
E-mail: jinguangyin18@nudt.edu.cn, huangjincai@nudt.edu.cn\protect\\
\IEEEcompsocthanksitem $^{*}$Both authors contributed equally to this research.
}}

\markboth{}
{Shell \MakeLowercase{\textit{et al.}}: Bare Demo of IEEEtran.cls for Computer Society Journals}

\IEEEtitleabstractindextext{%
\begin{abstract}
Travel time estimation is one of the core tasks for the development of intelligent transportation systems.
Most previous works model the road segments or intersections separately by learning their spatio-temporal characteristics to estimate travel time. However, due to the continuous alternations of the road segments and intersections in a path, the dynamic features are supposed to be coupled and interactive. Therefore, modeling one of them limits further improvement in accuracy of estimating travel time.
To address the above problems, a novel graph-based deep learning framework for travel time estimation is proposed in this paper, namely Spatio-Temporal Dual Graph Neural Networks (STDGNN). Specifically, we first establish the node-wise and edge-wise graphs to respectively characterize the adjacency relations of intersections and that of road segments.
In order to extract the joint spatio-temporal correlations of the intersections and road segments, we adopt the spatio-temporal dual graph learning approach that incorporates multiple spatial-temporal dual graph learning modules with multi-scale network architectures.
Finally, we employ the multi-task learning approach to estimate the travel time of a given whole route, each road segment and intersection simultaneously.
We conduct extensive experiments to evaluate our proposed model on three real-world trajectory datasets, and the experimental results show that STDGNN significantly outperforms several state-of-art baselines.

\end{abstract}

}

\maketitle

\IEEEdisplaynontitleabstractindextext

\IEEEpeerreviewmaketitle


\section{Introduction}
\IEEEPARstart{W}{ith} the development of modern urbanization, more powerful sensing devices and application terminals can collect more traveling trajectory data. It is conducive to fully exploit these data for real-time monitoring and prediction of traffic dynamics in urban area, thereby promoting the construction of smart cities. Travel time estimation (TTE) is one of the most promising trajectory data mining task, which is widely used in routing planing, navigation and ride sharing. The function of TTE has been deployed in many online map service providers such as Google map and AutoNavi to help users plan routes intelligently in advance. For complex spatio-temporal dynamics in urban transportation systems, how to estimate the travel time accurately has become a critical concern for these map service providers in improving practical value of their applications. 

Since a path is composed of road segments and intersections alternately, the latent features of both the intersections and road segments are pivotal in travel time estimation. Specifically,  for the intersections, traffic signals are introduced to control traffic movement to maintain the traffic order, which have significant influence on the speed of traffic flow. According to~\cite{tirachini2013estimation}, due to the traffic control and congestion, stopping at intersections caused at least 10\% of travel time delays, which means that the impacts of intersections should be taken seriously. Meanwhile, the traffic speed attached to the road segments has a more direct impact on travel time estimation. However, most previous works only exploit the features of either road segments~\cite{wang2019simple,fu2020compacteta,fang2020constgat} or intersections~\cite{wang2019learning} for estimating the travel time. Obviously, the main limitation of these works is that they do not take the features of 
both intersections and road segments into account simultaneously, which prevents further improvement in accuracy of estimation. 

However, it is challenging to jointly model the intersections and road segments for capturing more comprehensive dynamic information in road networks. First, for both intersections and road segments, they have their own independent static attributes that could affect the traffic state such as distances of the roads and traffic lights at intersections. Second, there exist some complex and interactive relations between the intersections and road segments. To be specific, since road segments and intersections are alternately connected in the road networks, the state of traffic flow can propagate in such alternating sequence. If congestion occurs at a certain intersection, the traffic state of the neighboring intersections and the multiple directly connected road segments could be affected first~\cite{chen2020multi}.
Third, there are complex spatial and temporal properties along with their complex relations. On the one hand, due to the interaction between intersections and road segments, the spatial properties of both of them could be closely related with each other in a large scale of regions.
On the other hand, the spatial and temporal properties cannot be learned separately, since the spatial properties are commonly time varying in road networks. For instance, the traffic state of downtown region could be impeded in the rush hours, but it could be relatively smooth in the non-rush hours.  

To tackle these problems, we present a novel dual graph deep learning framework STDGNN. First, we construct node-wise and edge-wise graphs to respectively characterize the independent structural features of the intersections and road segments. Then we design a spatio-temporal dual graph learning approach to capture the joint correlations of the dual graph. Specifically, we adopt spatial graph convolution networks and temporal convolution networks to capture the independent spatio-temporal correlations of the intersections and road segments. In order to further learn the interactive relations between the node-wise and edge-wise graph, we propose a dual graph interaction mechanism to aggregate the latent representations from them. In addition, we also present a novel multi-scale architecture to fully exploit multi-level latent information.
Finally, we adopt a multi-task learning approach to estimate the travel time of the paths, road segments and intersections at the same time. 


We summarize our main contributions as follows:

(1) We propose an end-to-end spatio-temporal graph deep learning model for multi-task travel time estimation. As far as we know, it is first attempt to jointly model the intersections and road segments based on dual graph learning for travel time estimation. 
 
(2) We build node-wise and edge-wise graphs to respectively characterize the features of both intersections and road segments and present a spatio-temporal dual graph learning layer to capture the independent and interactive latent correlations from them. 
Further, we propose a novel multi-scale architecture to fully exploit multi-level spatio-temporal information to improve the quality of the final representations from spatio-temporal learning module.
 
(3) We conduct extensive experiments to evaluate the performance of our proposed model on three different datasets. From the experimental results, we can observe that STDGNN outperforms than other state-of-art baseline under multiple metrics. 

The rest of this paper is organized as follows. We first review the related works and propose a comprehensive classification in Section 2. Then the travel time estimation problem is formulated in Section 3. Motivated by the challenges, we introduce the details of our solutions in Section 4. After that, we design multiple experiments to evaluate our model in Section 5, where the ablation studies and parameter studies are conducted. Finally, we conclude our paper in Section 6. 
\section{Literature Review}
In this section, we introduce several related works on travel time estimation from two perspectives, traditional methods and deep learning methods respectively. 

\subsection{Traditional Methods for Travel Time Estimation}
There exits a large body of traditional methods on travel time estimation. These previous works can be roughly classified into three categories: road segment-based, path-based and statistical learning-based methods.  

\textbf{Road Segment-based Methods.} In the early stage, there are many individual road segment-based works~\cite{rice2004simple,wu2004travel,sevlian2010travel,jenelius2013travel,nath2010modified} for travel time estimation. For example, Wu et al.~\cite{wu2004travel} adopt support vector regression to predict travel time of road segments. Nath et al.~\cite{nath2010modified} propose a modified K-means clustering approach for historical traffic data and estimate travel time of each road segment based on different clusters. The road segment-based methods can hardly achieve accurate estimation because they ignore the intersections and sequential correlations among road segments. 

\textbf{Path-based Methods.} Some path-based methods~\cite{yuan2011t,luo2013finding,rahmani2013route,wang2014travel,yang2018pace,wang2019simple} are introduced to take the sequential information of the whole paths into account. For example, Wang et al.~\cite{wang2019simple} aggregate the paths with adjacent origination and destination to estimate the travel time of the whole paths. Yu et al.~\cite{wang2014travel} integrate multiple spatio-temporal information to construct three-dimensional tensor and then estimate travel time of paths by dynamic programming solution. Although these methods take more comprehensive information into consideration, they rely on the prior knowledge modeling without data-driven mechanism. 

\textbf{Statistical Learning-based Methods.} Some state-of-art statistical learning-based methods are presented to improve the estimated accuracy by the data-driven manner. Zhang et al.~\cite{zhang2015gradient} propose a gradient boosting tree regression methods to achieve a higher accuracy for highway travel time estimation.  Gupta et al.~\cite{gupta2018taxi} propose a ensemble learning methods to take advantage of multiple variants of gradient boosting models for taxi traval time prediction. 
However, the bottleneck of these methods is that they cannot effectively capture the complex and dynamic spatio-temporal correlations from the given path to obtain higher accuracy. 

\subsection{Deep learning Methods for Travel Time Estimation}
Recently, deep learning-based methods become increasingly important in travel time estimation~\cite{wang2018will,wang2018learning,zhang2018deeptravel,fu2019deepist}. These related works can be divided into two groups, classical deep learning-based methods and graph deep learning-based methods. 

\textbf{Classical Deep Learning-based Methods.} Classical deep learning-based methods mainly adopt variants of some classic deep learning models such as Convolution Neural Networks (CNNs) and Recurrent Neural Networks (RNNs). Wang et al.~\cite{wang2018will} first propose an efficiency deep learning framework that can capture spatial and temporal dependencies along the given path. Both of DeepTravel~\cite{zhang2018deeptravel} and DeepIST~\cite{fu2019deepist} model the road network as a grid map, and then design deep framework to capture spatial and temporal patterns from the grid map for travel time estimation. Wang et al.~\cite{wang2018learning} propose a novel deep-wide framework to integrate multiple information efficiently for travel time estimation. However, although these classic deep learning methods capture the spatio-temporal correlations effectively, they still ignore the spatial structures of road networks in real-world.

\textbf{Graph Deep Learning-based Methods.} Graph Neural Network (GNN) is a fruitful tool for learning representation of structural data, which is widely applied in natural language processing, computer vision and social networks analysis~\cite{wu2020comprehensive}. In recent years, the family of GNN has been expanded rapidly and a new branch named Spatial-Temporal Graph Neural Network (STGNN) has begun to exist. STGNN is a framework that integrates traditional GNN and temporal learning modules, which can capture spatial and temporal correlations simultaneously for non-euclidean data. Most existing works adopt STGNN in intelligent transportation~\cite{he2018stann,zhang2018gaan,yu20193d,wang2020traffic,chen2020multi}, environment monitoring~\cite{jin2021gsen,wang2020pm2}, social computing~\cite{wang2020deep,yu2020spatio} and so on. Among them, the achievement of STGNN in traffic prediction is the most attractive. Yu et al.~\cite{yu2017spatio} first propose an end-to-end STGNN framework for traffic prediction, which captures the spatial and temporal correlations respectively by GNN and 1D Convolution Neural Network. Li et al.~\cite{li2017diffusion} first combines the structure of Seq2seq with diffusion GNN for multi-step flow prediction. Guo et al.~\cite{guo2019attention} proposes an attention-based STGNN framework to capture superior long-range temporal correlations. There are also many improved variants based on these works. To capture multi-source spatial dependencies, geng et al.~\cite{geng2019spatiotemporal} and Jin et al.~\cite{jin2020urban} use multi-graph modeling approach to enhance the spatial latent representation. To capture spatio-temporal correlations comprehensive and efficient, both~\cite{wu2019graph} and~\cite{wu2020connecting} involve adaptive graph and dilated temporal convolution with multiple skip connections. 
To capture the non-euclidean structural information from road networks in real-world, graph neural networks are also adopted in travel time estimation~\cite{fang2020constgat,fu2020compacteta}. For example, ConSTGAT~\cite{fang2020constgat} involves a graph attention mechanism for learning spatial and temporal information attached to the road neworks, and then capture the contextual information of the given path by simple convolutions. Fu et al.~\cite{fu2020compacteta} design an online system to estimate the travel time of road segments and whole paths by applying a graph attention network.  
However, these works only consider the spatio-temporal attributes of road segments but ignore the interactive correlations between intersections and road segments. Different from these works, we propose a novel dual graph neural network to jointly model the intersections and road segments.


\section {Preliminary}

In this section, we introduce some key preliminaries and then present the research problem.

\textbf{Trajectory}: A trajectory $Y$ can be treated as a series of sequential continuous GPS points $Y=\{y_1,y_2,...,y_{|Y|}\}$, each of which contains three different attributes: the location (latitude ($y_i^{l_1}$) and longitude ($y_i^{l_2}$)) and the timestamp ($y_i^t$) respectively.

\textbf{Road Network}: The connection of road network can be seen as a natural  directed graph $G=(V, E)$. Notation $V$ and $E$ represent the node set and edge set. Node $v\in V$ is an intersection while edge $e_{i,j}\in E$ is a road segment that is adjacent to intersection $v_i$ and $v_j$.  For convenience, the meanings of road segments and links are equivalent in the following paper. 

\textbf{Path}: Path $P$ can be treated as a sequence of of alternating links and intersections. We have $P = \{e_{o,1}, v_1, e_{1,2}, v_2, ..., e_{M-1,d}\}$, where $e_{o,2}$ and $e_{M-1,d}$ may be partial link.

\textbf{Objective}: With the given path $P$ and departure time $t_d$, the aim is to forecast the travel time $t_{\Theta}$ based on historical trajectory dataset $D$ and corresponding road network $G$. We need to design a no-linear function $\mathcal{F}(\cdot)$ for travel time estimation, which is defined as: 
\begin{equation}
t_{\Theta}\leftarrow \mathcal{F}(P,t_d|D,G), 
\end{equation}
where $\Theta=\{P,v_P,s_P\}$ denotes the set of the entire paths and and the continuous roads and intersections on them. 

\section{Methodology}
\begin{figure}[tb]
 \centering
 \includegraphics[width=0.47 \textwidth]{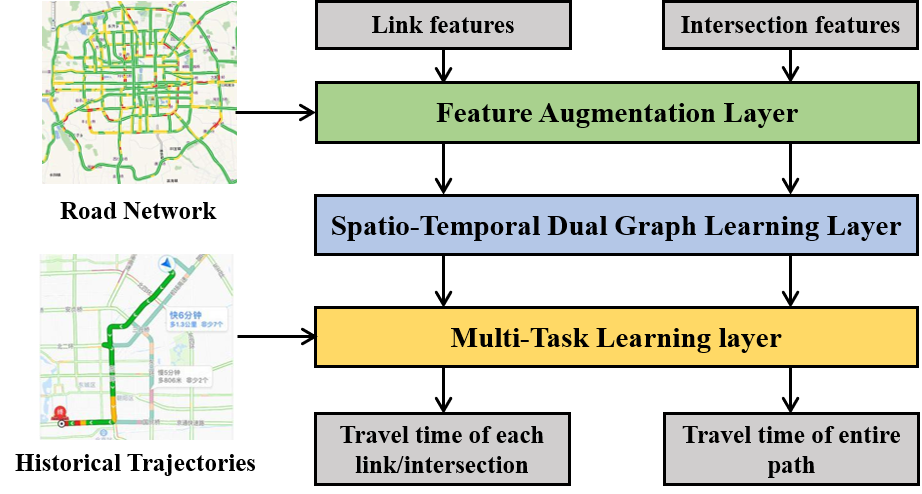}
 \caption{The overview of STDGNN.}
 \label{fig:model} 
 \end{figure}

To achieve a higher accuracy in travel time estimation, fully exploiting the latent features of intersections and road segments is a promising direction. However, jointly modeling the intersections and road segments is a challenging problem. 
First, both intersections and road segments have their independent complex spatio-temporal dynamic features. Second, the dynamics of these two kinds of features are not independent and there are some complex interactions between them. To tackle these limitations, we present a novel framework STDGCN, which is composed of three main layers: feature augmentation layer, spatio-temporal dual graph learning layer and multi-task learning layer. The overview of our proposed model is designed as shown in Fig.~\ref{fig:model}.
 

 \begin{figure*}[h]
 \centering
 \includegraphics[width=0.9 \textwidth]{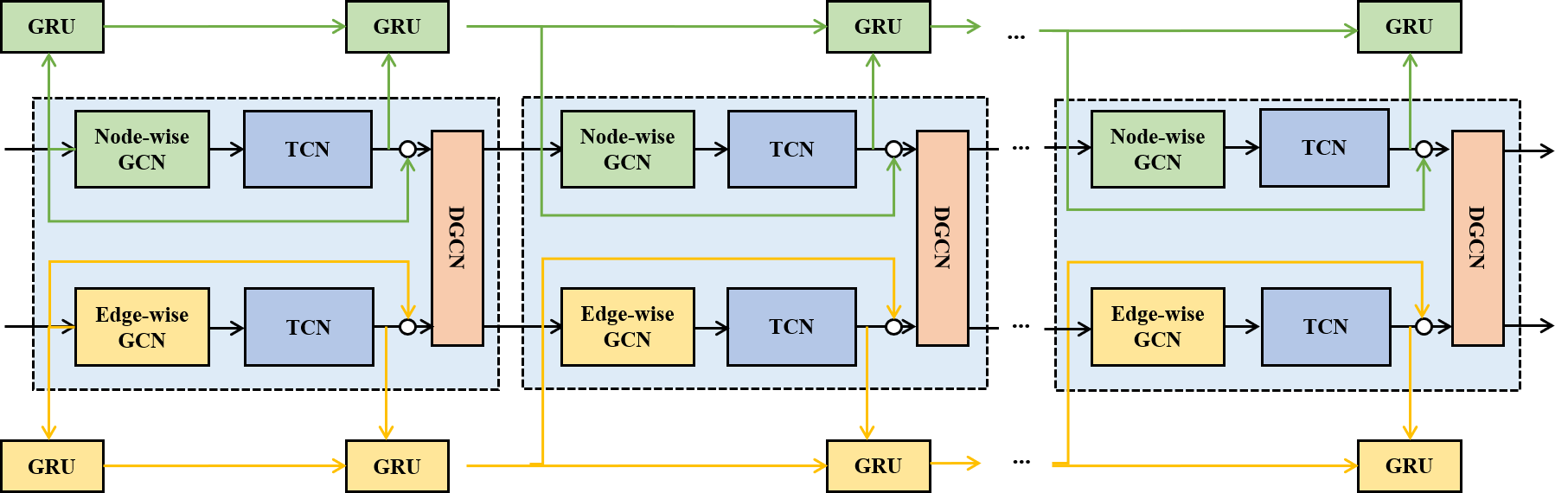}
 \caption{The overview of the spatio-temporal dual graph learning layer. This layer consists of multiple spatio-temporal dual graph learning cells, which is is enclosed by the blue dashed block diagrams. Each cell is composed of node-wise GCN, edge-wise GCN, temporal convolution network (TCN) and dual graph convolution network (DGCN).  In addition, a novel gated recurrent unit (GRU) based multi-scale architecture is adopted among different cells. }
 \label{fig:spatio_temporal_graph_layer} 
 \end{figure*} 
 
\begin{itemize}
\item \textbf{Feature Augmentation Layer.} The enhanced representations of the links and intersections provide a better starting point to efficiently learn complex features for other layers of the framework. 
We realize the initialization of the representations by integrating some external features and mapping them to the high-dimensional feature space.   
Especially for the intersections, we also consider the information from their adjacent links and aggregate them into the intersection representations.

\item \textbf{Spatio-temporal Dual Graph Learning Layer.} In order to jointly model the spatio-temporal dynamics of intersections and links, we first adopt the node-wise and edge-wise spatio-temporal graph convolutions to capture their independent features. Next, a dual graph interaction mechanism is introduced to aggregate the entangled features between intersections and links. 
Then, to fully exploit the multi-scale information, a novel multi-scale architecture is proposed to integrate the features from multiple scales of spatio-temporal dual graph learning cells. 

\item \textbf{Multi-task Learning Layer.} To estimate the travel time of each link, intersection, and the entire path at the same time, we adopt a multi-task learning approach to model the representations from alternating links and intersections. Meanwhile, a multi-task loss function is introduced to balance local (intersections and links) and global (the entire path) estimations.
 \end{itemize}

 \subsection{Feature Augmentation Layer}
 
 
Since some important features attached to the links could affect the travel time of a path (e.g., the links' speed, type and distance), 
we initialize the link representations by concentrating these features and mapping them into high-dimensional latent space. To be specific, the link representations at time step $t$ can be defined as follows: 
\begin{equation}
\boldsymbol{h}_e(t)={\rm{tanh}}(\boldsymbol{W}_a\cdot [s_e(t), d_e, p_e]),
\end{equation}
where $s_e(t)$ is the links' mean speed at time $t$, $d_e$ is the links' distance, $p_e$ is the links' type. $\boldsymbol{W}_a$ is a learnable weight to map the link features into high-dimensional latent space and $[\cdot]$ denotes the concatenation operator.
 
For intersections, since any of them are adjacent to multiple links, the representations of them can be initialized by aggregating the representations of the adjacent links. 
Meanwhile, there are also some specific geo-spatial features attached to the intersections, such as traffic lights and special signs. Hence, we define the intersection representations at time $t$ as follows:
\begin{equation}
\boldsymbol{z}_v(t)={\rm{tanh}}(\boldsymbol{W}_b\cdot[\sum_{i\in F(v)}{\boldsymbol{h}_i(t)}, p_v]),
\end{equation}
where $F(v)$ is the set of links connecting with intersection $v$. $p_v$ denotes the geo-spatial features of intersections. $\boldsymbol{W}_b$ is the learnable parameter to map the intersection features into high-dimensional latent space.

In this way, both the latent representations of links and intersections can have sufficient capabilities to cover the complex spatio-temporal dynamics by feature augmentation.

\subsection{Spatio-temporal Dual Graph Learning Layer}   
The core of STDGNN is the spatio-temporal dual graph learning layer, as shown in Fig.~\ref{fig:spatio_temporal_graph_layer}. The function of this layer is to obtain the latent representation of road segments and intersections for next time interval based on previous $T$ time steps.
Then we introduce three important modules in this layer, including dual graph construction module, spatio-temporal dual graph learning module and the multi-scale architecture module, which can jointly capture the multi-scale spatio-temporal correlations from intersections and road segments.

\subsubsection{Dual Graph Construction Module} 
Existing graph deep learning based models exploit GCN with a weighted adjacency matrix to learn the correlations among the links, while the intersections connecting the road segments are not considered~\cite{fang2020constgat,fu2020compacteta}. This neglects not only the independent features of intersections but also the complex interactive correlations between the links and their connected intersections. To fully exploit the joint latent features of the links and intersections, we present a spatio-temporal dual graph convolution approach to model their complicated correlations synchronously. 

First we need to establish node-wise graph and edge-wise graph to characterize the structural features of the intersections and links respectively.

Let $G_n =(V_n,E_n,\boldsymbol{W_n})$ denotes the node-wise graph, whose nodes $V_n$ and edges $E_n$ respectively represent intersections and the links respectively. Let $G_e =(V_e,E_e,\boldsymbol{W_e})$ denotes the corresponding edge-wise graph, and the nodes $V_e$ of edge-wise graph are the ordered edges in $E_n$ of the node-wise graph, i.e., $V_e={(i\rightarrow j),(i,j)\in E_e}$ and $|V_e|=|E_n|$. $\boldsymbol{W_n}$ and $\boldsymbol{W_e}$ are the weighted adjacency matrices to characterize the adjacent relations of intersections and links respectively.

Node-wise graph $G_n$ is a a directed weighted graph, and the basic connection between different nodes is determined by the spatial structures of road networks. However, the connection weights between any two spatially adjacent nodes $i$ and $j$ are different. 
Intuitively, if both node $i$ and node $j$ have larger out-degrees and in-degrees, this means traffic flow in the same direction could be shared by more intersections. Thus, the connection weight between node $i$ and node $j$ is relative small. The edge weights $\boldsymbol{W_n}$ of node-wise graph can be computed as follows:
\begin{equation}
w_{n,i:j} = R_{ij} exp(-\frac{(d^+(i)+d^-(j)-2)^2}{\sigma^2}),i\ne j,
\end{equation}
where $d^+(i)$ and $d^-(j)$ represents the number of out-degree of node $i$ and in-degree of node $j$, respectively. $\sigma$ denotes the standard deviation of node degrees. $R_{ij}$ represents the spatially adjacent relation between node $i$ and node $j$ in the road network. Specifically, for each pair $(i,j)\in V_n\times V_n$, $R_{ij}$ is 1 if $(i,j)\in E_n$ and 0 otherwise.

Edge-wise graph $G_e$ is also a directed weighted graph and the connection weight $w_{e,i:j}$ between different links is defined as the volume of the traffic. Obviously, if there exists larger traffic volume from link $i$ to link $j$ than that from link $i$ to other neighboring links, the connection between link $i$ to link $j$ is stronger.
Thus, the edge weights $\boldsymbol{W_e}$ can be computed as follows:
\begin{equation}
w_{e,i:j} = R_{ij} \frac{z_{ij}}{\sum_{k\in N(i)} z_{ik}},i\ne j,
\end{equation}
where $z_{ij}$ represents the number of trajectories traversed from link $i$ to link $j$. $N(i)$ denotes the 1-hop neighbors of node $i$.

\begin{figure}[h]
\vspace{-2mm}
\centering
\includegraphics[width=0.48 \textwidth]{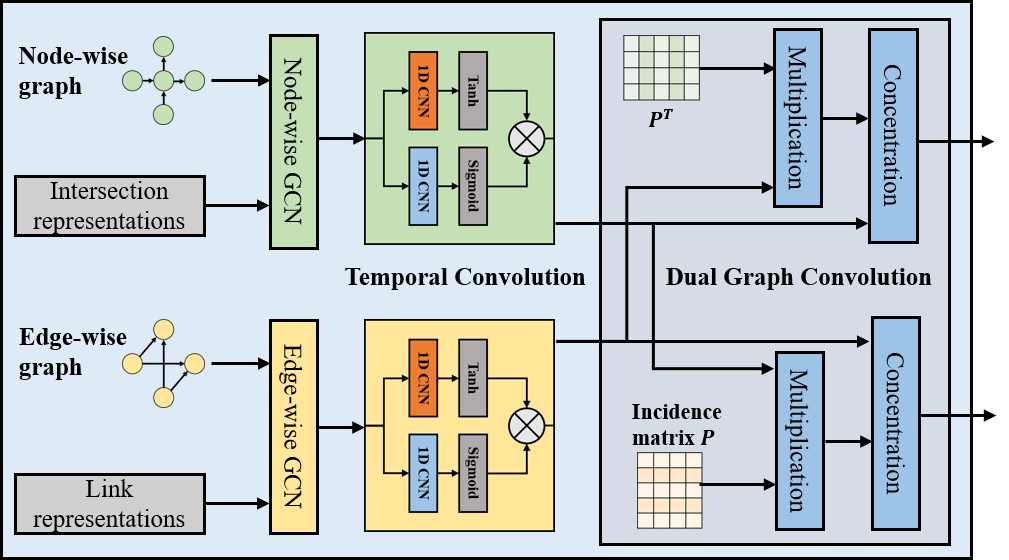}
\caption{The detailed architecture of Spatio-temporal Dual Graph learning Module. }
\label{fig:cell} 
\vspace{-2mm}
\end{figure}

\subsubsection{Spatio-temporal Dual Graph Learning Module} 
Since both links and intersections have their independent spatio-temproal latent correlations, we design an architecture of alternating graph convolution network (GCN) and temporal convolution network (TCN) to capture spatio-temporal dependencies from node-wise graph and edge-wise graph respectively, as shown in Fig.~\ref{fig:cell}. 

To capture the spatial dependencies, we select the simple graph convolution approach to aggregate information of 1-hop neighbors~\cite{kipf2016semi}. The graph convolution operator is formulated as: 
\begin{equation}
\boldsymbol{Z}^{(l+1)}=\sigma(\boldsymbol{L}\cdot\boldsymbol{Z}^{(l)}\cdot \boldsymbol{W}^{(l)})
\end{equation}
where $\boldsymbol{L}=\widetilde{\boldsymbol{D}}^{-1/2}\widetilde{\boldsymbol{A}}\widetilde{\boldsymbol{D}}^{-1/2}$, $\widetilde{\boldsymbol{A}}=\boldsymbol{A}+\boldsymbol{I}$, $\widetilde{\boldsymbol{D}}=diag(\sum_j\widetilde{A}_{1j},...,\sum_j\widetilde{A}_{Nj})$. $\boldsymbol{A}$ is a symmetric adjacency matrix, and $\boldsymbol{I}$ is an identity matrix. $\sigma$ denotes a non-linear activation function, and we apply function \emph{ReLU} for graph convolution. $\boldsymbol{Z}^{(l)}$ denotes the input graph representation and $\boldsymbol{Z}^{(l+1)}$ denotes the output graph representation.

However, the node-wise graph and edge-wise graph are both directed weighted graphs. Thus, we adopt a dual directional graph convolution to make full of directed graphs, which is defined as:
\begin{equation}
\label{eq:gcn_out}
\boldsymbol{Z}^{(l+1)} = \sigma(\boldsymbol{L}\cdot\boldsymbol{Z}^{(l)}\cdot \boldsymbol{W}^{(l)} +\boldsymbol{L}^{T}\cdot\boldsymbol{Z}^{(l)}\cdot \boldsymbol{W}^{(l)})
\end{equation}

Then, we use temporal convolution network to capture temporal dependency of both intersections and links respectively, as shown in Fig.~\ref{fig:cell}. We adopt gated mechanism into the temporal convolution network, which is proven as an effective approach to control the reserved information~\cite{shi2016end}. The temporal convolution operation is formulated as follows:
\begin{equation}\label{eq:tcn}
\boldsymbol{Z_{T}}=\Vert_{0}^{N} \sigma_{1}(\boldsymbol{z_{i}}\star\Gamma_{1}(\boldsymbol{\theta_{1}}))\odot\sigma_{2}(\boldsymbol{z_{i}}\star\Gamma_{2}(\boldsymbol{\theta_{2}}))
\end{equation}
where $\boldsymbol{\theta_{1}}$, $\boldsymbol{\theta_{2}}$ are the learnable parameters of temporal convolution, $\odot$ is element-wise product operation, $\sigma_{1}(\cdot)$ and $\sigma_{2}(\cdot)$ are activation functions of two different temporal convolution models respectively. Empirically, \emph{Tanh} function can be selected as $\sigma_{1}(\cdot)$ and \emph{Sigmoid} function is usually selected as $\sigma_{2}(\cdot)$ to control the ratio of information passed. $\boldsymbol{z_{i}}$ represents the latent representation of node $i$. The temporal convolution network is applied to each node in node-wise graph or edge-wise graph, thus the output is the result of integrating all $N$ nodes of node-wise graph or edge-wise graph. In addition, in order to maintain the unity of the time dimension in spatio-temporal dual graph learning layer, we adopt pre-padding approach for temporal convolution network.  

After spatio-temporal convolution for node-wise graph and edge-wise graph, we obtain the independent spatio-temporal latent representation of the two graphs. In order to capture the latent interaction patterns between links and intersections, we propose a dual graph convolution mechanism. 
Thus, the dual graph convolution is formulated as:
\begin{equation} \label{eq:dgcn}
\begin{split}
\boldsymbol{Z_n}^{(l+1)}&=\theta_{n*G}^{(l)}
[\boldsymbol{Z_n}^{(l)},\boldsymbol{P}\boldsymbol{Z_e}^{(l)}]    \\
\boldsymbol{Z_e}^{(l+1)}&=\theta_{e*G}^{(l)} [\boldsymbol{Z_e}^{(l)},\boldsymbol{P}^T\boldsymbol{Z_n}^{(l+1)}]
\end{split}
\end{equation}
where $\theta_{n*G}$ and $\theta_{e*G}$ are respectively the node-wise and edge-wise graph convolution operation, $\boldsymbol{Z_n}$ and $\boldsymbol{Z_e}$ are respectively the node-wise and edge-wise latent representation, and $\boldsymbol{P}\in \mathbb R^{|V_n|\times |E_n|}$ is the incidence matrix that encodes the connections between nodes and edges, defined as: $P_{i,(i\rightarrow j)}=P_{j,({i\rightarrow j})}=1$ and 0 otherwise.

To respectively obtain the latent representations of node-wise graph and edge-wise graph for the next time interval, we adopt two fully connected layers after the last dual graph convolution operation in spatio-temporal dual graph learning layer, which can be expressed as:
\begin{equation}
\begin{split}
\boldsymbol{Z_n(t)}&=(\boldsymbol{W}_n\cdot \boldsymbol{Z_n}^{(l+1)}),\\
\boldsymbol{Z_e(t)}&=(\boldsymbol{W}_e\cdot \boldsymbol{Z_e}^{(l+1)})
\end{split}
\end{equation}
We assume that $\boldsymbol{Z_n}^{(l+1)}$ and $\boldsymbol{Z_e}^{(l+1)}$ are respectively the output from the last dual convolution operation. $\boldsymbol{W}_n$ and $\boldsymbol{W}_e$ are respectively the weight parameters of the two fully connected layer for node-wise representations and edge-wise representations. $\boldsymbol{Z_n(t)}$ and $\boldsymbol{Z_e(t)}$ are the estimated latent representations of the dual graph in next time interval.

\subsubsection{Multi-scale Architecture Module} 
By stacking multiple spatio-temporal dual graph learning cells, we can expand the receptive spatial and temporal range. However, as the network goes deeper, each node of node-wise graph and edge-wise graph tend to have the global spatio-temporal dependencies, while losing their local spatio-temporal dependencies. Therefore, to better capture both global and local spatio-temporal dependencies, we make full of the multi-scale information.
There are some previous techniques to reserve the local information in deep networks such as residual connection~\cite{szegedy2017inception} and skip connection~\cite{mao2016image}, but these methods do not fully exploit the multi-level information in shallow layers.
In~\cite{Luan2019Break}, the snowball approach was first proposed to incrementally concatenate multi-scale features for graph convolution. Motivated by this work, we can adopt the similar approach to capture spatio-temporal features from outputs in shallow layers. But different from~\cite{Luan2019Break}, we use gated recurrent units (GRU)~\cite{Chung2014Empirical} to select important information from previous layers, since GRU can be a simple but effective filter for sequential information, as shown in Fig.~\ref{fig:spatio_temporal_graph_layer}. The process of computation can be formulated as follows:
\begin{equation}
\vspace{-2mm}
\begin{split}
& \boldsymbol{c_n}^{(0)}={\rm G_n}(\boldsymbol{r_n}^{(0)},\boldsymbol{c_n}^{(-1)}),\\
&\boldsymbol{c_e}^{(0)}={\rm G_e}(\boldsymbol{r_e}^{(0)},\boldsymbol{c_e}^{(-1)}),\\
& \boldsymbol{r_n}^{(1)}={\rm T_n}({\rm S_n}(\boldsymbol{r_n}^{(0)})), \\ 
& \boldsymbol{r_e}^{(1)}={\rm T_e}({\rm S_e}(\boldsymbol{r_e}^{(0)})),\\ 
& \boldsymbol{r_n}^{(2)}={\rm T_n}({\rm S_n}({\rm D}([\boldsymbol{r_n}^{(0)},\boldsymbol{r_n}^{(1)}],\boldsymbol{r_e}^{(1)})),\\
& \boldsymbol{r_e}^{(2)}={\rm T_e}({\rm S_e}({\rm D}([\boldsymbol{r_e}^{(0)},\boldsymbol{r_e}^{(1)}],\boldsymbol{r_n}^{(1)})),\\
& \boldsymbol{r_n}^{(l+1)}={\rm T_n}({\rm S_n}({\rm D}([{\rm G_n}(\boldsymbol{r_n}^{(l-1)},\boldsymbol{c_n}^{(l-2)}), \boldsymbol{r_n}^{(l)}],\boldsymbol{r_e}^{(l)}))),\\
& \boldsymbol{r_e}^{(l+1)}={\rm T_e}({\rm S_e}({\rm D}([{\rm G_e}(\boldsymbol{r_e}^{(l-1)},\boldsymbol{c_e}^{(l-2)}), \boldsymbol{r_e}^{(l)}],\boldsymbol{r_n}^{(l)}))),\\
& l = 2, 3, 4, \dots
\end{split}  
\vspace{-2mm}
\end{equation}\label{eq:gcn_gru}

$T_n(\cdot)$ and $T_e(\cdot)$ are respectively the temporal convolution function of node-wise and edge-wise latent representation, which refers to Eq.~\ref{eq:tcn}.
$S_n(\cdot)$ and $S_e(\cdot)$ are respectively the graph convolution function of node-wise and edge-wise graph, which refers to Eq.~\ref{eq:gcn_out}.
$D(\cdot)$ denotes the dual graph convolution function, which refers to Eq.~\ref{eq:dgcn}.
$\boldsymbol{r_n}^{(l)}$ and $\boldsymbol{r_e}^{(l)}$ are respectively the latent representation of intersection and link at $l$-th layer.
$\boldsymbol{c_n}^{(l)}$ and $\boldsymbol{c_e}^{(l)}$ are respectively the hidden state of node-wise graph and edge-wise graph after we processed the information from $(l-1)$-th layer, and its initial value $\boldsymbol{c_n}^{(-1)}=\boldsymbol{0}$ and $\boldsymbol{c_e}^{(-1)}=\boldsymbol{0}$. 
In this architecture, we adopt two different GRU to process the information from the two different latent representation. $G_n(\cdot)$ and $G_e(\cdot)$ are respectively the GRU function for node-wise and edge-wise latent representation. In mathematical form, the two different GRU are the same, which is defined as follows:
\begin{equation}
\begin{split}
 &  u=\sigma(\boldsymbol{W}_{u_1} \boldsymbol{r}(t)+\boldsymbol{U}_{u_1} {\boldsymbol{c}}(t-1)+\boldsymbol{b}_{u_1}),\\ 
&  x=\sigma(\boldsymbol{W}_{x_1} \boldsymbol{r}(t)+\boldsymbol{U}_{x_1} {\boldsymbol{c}}(t-1)+\boldsymbol{b}_{x_1}),\\
&   \boldsymbol{c}'(t)= {\rm tanh}(\boldsymbol{W}_{h_1} \boldsymbol{r}(t)+\boldsymbol{U}_{h_1}(x\odot {\boldsymbol{c}}(t-1))+\boldsymbol{b}_{h_1}),\\
& \boldsymbol{c}(t)=u\odot \boldsymbol{r}(t-1)+(1-u)\odot \boldsymbol{c}'(t),
\end{split}      
\end{equation}
where $\odot$ is the element-wise multiplication. $\boldsymbol{W}_{u_1},\boldsymbol{U}_{u_1},\boldsymbol{W}_{x_1},\boldsymbol{U}_{x_1},\boldsymbol{W}_{h_1},\boldsymbol{U}_{h_1}$ are the parameters to be learned. $\boldsymbol{b}_{u_1},\boldsymbol{b}_{x_1},\boldsymbol{b}_{h_1}$ are biases.
  
\subsection{Multi-task Learning Layer}
  
\begin{figure}[h]
\centering
\includegraphics[width=0.5 \textwidth]{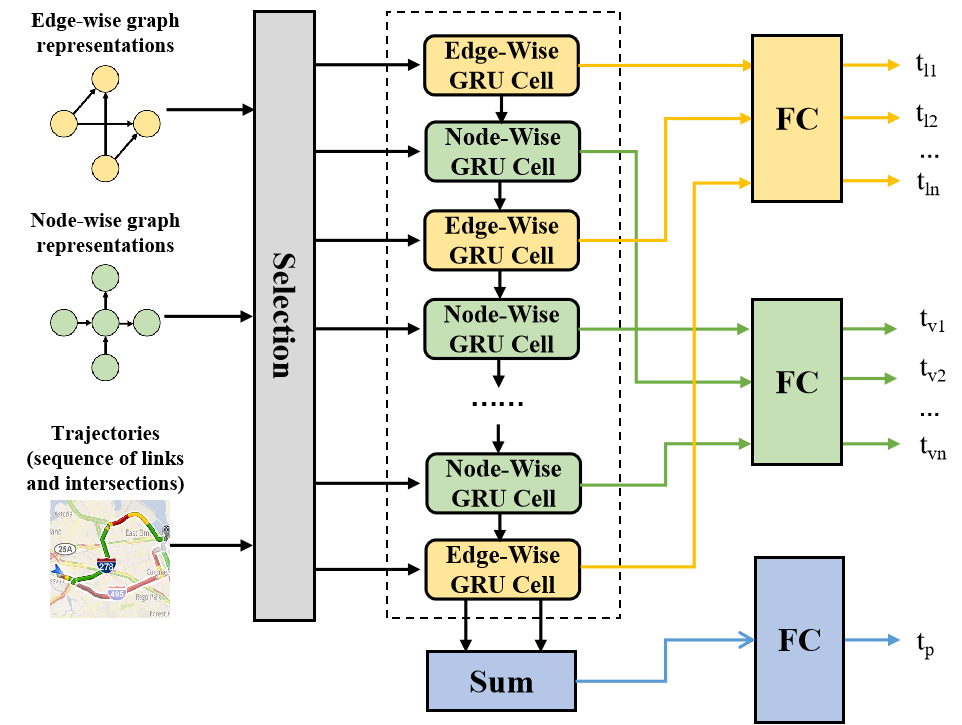}
\caption{An overview of multi-task learning layer. }
\label{fig:multitask_learning_layer} 
\end{figure}
 
Since the trajectories are always sequentially dependent, we use the sequential information from specific links and intersections to further capture the temporal dynamics hidden in the latent representations. As seen in Fig.~\ref{fig:multitask_learning_layer}, we select the corresponding node latent representation from node-wise graph and edge-wise graph based on the index of links and intersections passed by the query path. We also employ two different GRU cell named Node-wise GRU and Edge-wise GRU to distinguish the latent representations from node-wise and edge-wise graph. For simplicity, we use $\boldsymbol{q}$ to uniformly represent latent representations from the two different sources. Thus, the formula can be expressed as:

\begin{equation}
\begin{split}
& u=\sigma(\boldsymbol{W}_{u_2} \boldsymbol{q}_i(t)+ \boldsymbol{U}_{u_2} \boldsymbol{c}_{i-1}(t)+ \boldsymbol{b}_{u_2}),\\ 
& r=\sigma(\boldsymbol{W}_{r_2} \boldsymbol{q}_i(t)+ \boldsymbol{U}_{r_2} \boldsymbol{c}_{i-1}(t)+ \boldsymbol{b}_{r_2}), \\ 
&  \boldsymbol{c}_i^{'}(t)={\rm tanh}(\boldsymbol{W}_{h_2} \boldsymbol{q}_i(t)+\boldsymbol{U}_{h_2}(r\odot \boldsymbol{c}_{i-1}(t))+\boldsymbol{b}_{h_2}), \\
&  \boldsymbol{c}_i(t)=u\odot \boldsymbol{c}_{i-1}(t)+(1-u)\odot \boldsymbol{c}_i^{'}(t),
\end{split}
\end{equation}
where $\boldsymbol{c}_i(t)$ is the final representation of $i$-th link or intersection at time step $t$.
     
\textbf{Prediction}: For local travel time estimation, we adopt a two-layer fully-connected network with size $[\lambda, 1]$ to get the final output. We define $t_{l_i}$ ( $t_{v_j}$) as the local travel time of the $i$-th link ($j$-th intersection).
   
For global travel time estimation, we need to use sum pooling to aggregate the latent representations sequence $\{\boldsymbol{c}_i\}$ into a fixed length vector. 
\begin{equation}
\boldsymbol{g}_{P} = \sum_{i=1}^{|P|} \boldsymbol{c}_i
\end{equation}
where $|P|$ is the total number of links and intersections in Path $P$. 
     
Finally, we push $\boldsymbol{g}_{P}$ into another two-layer fully-connected networks with size $[\lambda, 1]$, and obtain the global estimation.
   
\textbf{Loss Function}: MAPE is chosen as the loss function for global travel time estimation, which is defined as follows:
\begin{equation}
L_{P} = \sum_{i=1}^{|D|} \frac{|t_i-\widehat{t_i}|}{\widehat{t_i}}, 
\end{equation}  
where $t_i$ denotes the global estimation of travel time, and $\widehat{t}_i$ denotes the real travel time for $i$-th path. 
    
For the link estimation, we also select MAPE as the loss function, and compute the the average loss of all local links as the final loss,
\begin{equation}
L_{l} = \sum_{P\in D} \sum_{j=1}^{(|P|+1)/2} \frac{|t_{l_j}-\widehat{t_{l_j}}|}{\widehat{t_{l_j}}+\epsilon}, 
\end{equation}   
where $\widehat{t}_{l_i}$ represents the real travel time for $i$-th link. Similar with~\cite{wang2018will}, we add a parameter $\epsilon$ to avoid exploded loss value when the denominator is too small. During the training phase, we set $\epsilon$ as 5.
    
For the intersection estimation, we also define the loss function as follows:
\begin{equation}
L_{v} = \sum_{P\in D} \sum_{j=1}^{(|P|-1)/2} \frac{|t_{v_j}-\widehat{t_{v_j}}|}{\widehat{t_{v_j}}+\epsilon}, 
\end{equation}   
where $\widehat{t}_{v_i}$ is the real travel time for $i$-th intersection.
    
The training objective is to minimize the combination of three terms, which is formulated as follows:
\begin{equation}
L = \alpha L_{P} + \beta L_{l} + (1-\alpha-\beta) L_{v},
\end{equation} 
$\alpha$ and $\beta$ are adjustable parameters to balance the three terms. In this case, we set $\alpha$ as 0.4 and $\beta$ as 0.3 by default.
    
Since our framework can perform global (entire path) and local (links and intersections) estimation of the travel time synchronously, it can be deployed for practical map navigation services.
\section{Experiment Settings and Results Analysis}
We conduct extensive experiments to evaluate our model on three real-word datasets. Our experiments aim to answer the following questions:

\noindent(1) How does our model perform compared with different baselines on three real-world datasets?

\noindent(2) How does our model perform compared with different baselines under different scenarios (e. g., rush hours and non-rush hours)?

\noindent(3) How each component affects the performance of our model in the ablation study? 

\noindent(4) How do some important hype-parameters (e. g., the number of spatio-temporal cells) influence the performance of our model?

\subsection{Experimental Settings}\label{sec:data_process}
We deploy STDGNN on three different real-world datasets, which are described as follows:

\begin{itemize}
\item \textbf{Chengdu dataset}: This dataset includes massive taxi trajectories of more than 14,000 taxis in 2014/08/01 to 2014/08/31 in Chengdu, China, which can be publicly available in~\cite{chengdu2014}.
\item \textbf{Porto dataset}: This dataset includes massive taxi trajectories of 442 taxis from 2013/07/01 to 2014/06/31 in Porto, Portugal, which can be publicly available from~\cite{porto2014}. 
\item \textbf{Beijing dataset}: This dataset contains 34,696 anonymous paths (without GPS point) during 8:00 AM - 1:00 PM on December 15, 2020 in Beijing, China.
\end{itemize}

\begin{figure}[h]
\centering
\includegraphics[width=0.45 \textwidth]{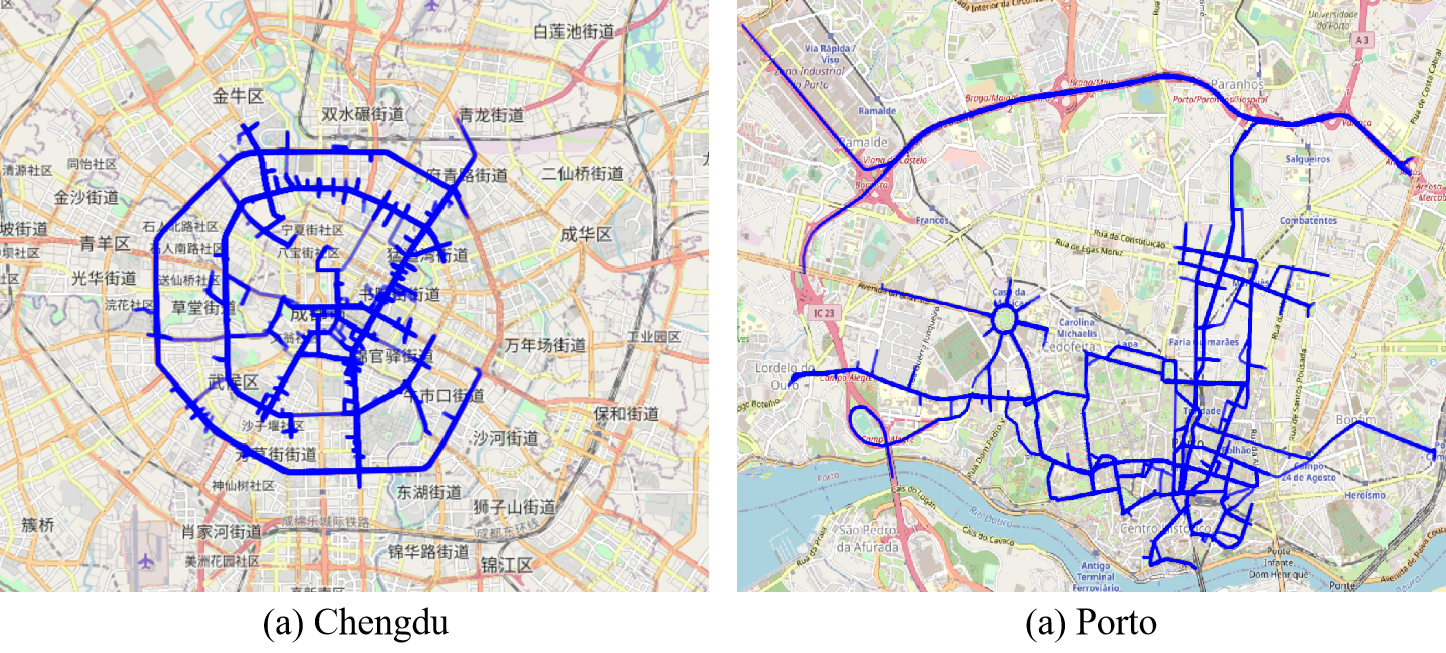}
\caption{The trajectories distributed in the downtown region of Chengdu and Porto.}
\label{fig:traj} 
\end{figure}

For two public datasets, we adopt OpenStreetMap~\cite{osm2014Steve} to access the structural road networks information of Chengdu and Porto, which contains road segments, intersections and adjacency relations of them. 
Next, for each dataset, the fast map matching algorithm~\cite{Can2018Fast} is adopted to map the trajectories into the actual road network to obtain the corresponding paths. Then, we define the paths that have significant deviation ($\pm 10\%$) from their raw trajectories as the mismatched ones to be filtered~\cite{fu2019deepist}.
Finally, 310,507 and 1,788,723 paths are collected from Porto dataset and Chengdu dataset respectively. However, we can not use all these paths since computation resources are limited .
Alternatively, we select the trajectories distributed in the downtown, for a relatively large number of trajectories traversed along the paths, as shown in Fig.~\ref{fig:traj}. Table~\ref{tab:stats} shows the numerical statistics of the three real-world datasets used in our experiments.

Since the some spatial and temporal factors could significantly influence the performance of the travel time estimation, we need to validate the generality of our model in different scenarios.
First, we select Beijing trajectories in three different scenarios, including the suburban areas, rush and non-rush hours of commercial areas.
For suburb scenario, since fewer trajectories are traversed along the paths in suburban areas, it can further evaluate how data sparsity affects the performance of our model.
For rush and non-rush scenarios, we define 7:00 - 9:00 AM and 5:00 - 7:00 PM at weekdays as the rush hours, and other time periods at weekdays and any time periods at weekends are defined as non-rush hours. Meanwhile, we define the scenarios of rush and non-rush hours in the urban areas. Table~\ref{tab:cdstats} shows some detailed information of Beijing dataset in these three scenarios.

\begin{table}[tb]
\centering
  \caption{The numerical statistics of three real-world datasets.} 
  \label{tab:stats}
  \scalebox{1.5}{
  \tiny
  \begin{tabular}{crrr}
	\toprule
			\textbf{Dataset}  & \multicolumn{1}{c}{\textbf{Chengdu}} & \multicolumn{1}{c}{\textbf{Porto}}  & \multicolumn{1}{c}{\textbf{Beijing}}\tabularnewline
		\midrule
		 \ number of trajectories & 15,303 & 12,683 & 34,696 \tabularnewline
		\ number of links & 873 & 544 & 714 \tabularnewline
		\ number of intersections & 807 & 450 & 320 \tabularnewline
		average travel time (s) & 246.54 & 248.49 & 343.3 \tabularnewline
		average moving distance (m) & 1435.34 & 1394.82 & 3929.7 \tabularnewline
				
		\bottomrule
  \end{tabular}}
\end{table}

\begin{table}[tb]
\centering
  \caption{The statistics of Beijing datasets in three scenarios, including the suburb, rush and non-rush hours.} 
  \label{tab:cdstats}
  \scalebox{1.4}{
  \tiny
  \begin{tabular}{crrrr}
	\toprule
			\textbf{Scenario}  & \multicolumn{1}{c}{\textbf{Suburb}} & \multicolumn{1}{c}{\textbf{Rush Hrs.}}  & \multicolumn{1}{c}{\textbf{Non-Rush Hrs.}}\tabularnewline
		\midrule
		 \ number of trajectories & 5,526 & 2,867  & 5,398\tabularnewline
		\ number of links & 227 & 250 & 250\tabularnewline
		\ number of intersections & 96 & 140 & 140\tabularnewline
		average travel time (s) & 586.85 & 332.77 & 275.11\tabularnewline
		average moving distance (m) & 5368.32 & 2738.17 & 2519.47\tabularnewline
				
		\bottomrule
  \end{tabular}}
\end{table}

\textbf{Evaluation Metric}. Similar to some previous works~\cite{wang2018will,fang2020constgat}, three metrics are selected to evaluate all methods: Mean Absolute Percentage Error (MAPE), Mean Average Error (MAE) and Root Mean Square Error (RMSE) respectively. In detail, both MAE and RMSE can measure the errors between the estiamtion and the ground truth, but RMSE is more sensitive for outlier errors while MAPE is a relative error to measure the estimation accuracy, which can eliminate the influence of the data scale. The lower these metrics, the better the performance of the models. The equation of them are formulated as follows:
  \begin{equation}
    {\rm RMSE}(\widehat{t}_{i},t_{i}) =\sqrt{\frac{1}{|D|}\sum_{i=1}^{|D|}{(t_{i}-\widehat{t}_i)^2}},
    \end{equation} 
   \begin{equation}
    {\rm MAE}(\widehat{t}_{i},t_{i}) =\frac{1}{|D|}\sum_{i=1}^{|D|}{|t_{i}-\widehat{t}_i|}, \quad\quad
    \end{equation} 
     \begin{equation}
    {\rm MAPE}(\widehat{t}_{i},t_{i}) =\frac{1}{|D|}\sum_{i=1}^{|D|}{\frac{|t_{i}-\widehat{t}_i|}{t_{i}}}.\quad
    \end{equation} 

For our model, the sizes of embedding dimension for intersections and links are set as 20. The number of spatio-temporal learning cell is set as 3 by default.
We use the historical information of intersection and link at first 12 time steps with slots of five minutes for temporal learning.
We adopt a two-layers fully-connected networks, and their hidden units are both set as 60. We train the model using Adam optimizer and the initial learning rate is set as 0.001 on all of three datasets. We deploy our model on TITAN Xp GPU with the implementation in Python with Pytorch 1.5, and repeat each experiment for three times. 
 
 \begin{table*}[htb]
\centering
  \caption{Performance of STDGCN and other baseline for global travel time estimation on three datasets. And the improvements under different metrics of STDGNN are compared with the sub-optimal methods, which is marked by the asterisk.} 
  \vspace{-3mm}
  \label{tab:overall_performance_path}
  \scalebox{0.9}{
  \begin{tabular}{c|ccc|ccc|ccc}
	\toprule
			  & \multicolumn{3}{c}{Chengdu} & \multicolumn{3}{c}{Porto} & \multicolumn{3}{c}{Beijing} \tabularnewline
			  Method & RMSE (sec) & MAE (sec) & MAPE & RMSE (sec)  & MAE (sec) & MAPE & RMSE (sec)  & MAE (sec) & MAPE\tabularnewline
		\midrule
		AVG & 202.43 & 132.86  & 0.6404 & 111.67 & 88.75  & 0.4509 & 338.43 &264.55  &0.7367\tabularnewline
		TEMP & 167.41 & 108.50  & 0.5538 & 98.01 & 78.12  & 0.4127 & - & - & - \tabularnewline
		GBDT & 165.03 & 101.68  & 0.5428 & 91.24 & 71.79  & 0.3914 & - & - & -\tabularnewline
		MlpTTE & 150.67 & 87.32  & 0.3076 & 59.46 & 44.49  & 0.1783 & - & - & -\tabularnewline
		RnnTTE & 155.16 & 89.37  & 0.3011 & 56.45 & 43.32  & 0.1702 & - & - & -\tabularnewline
		DeepTTE & 148.30* & 86.40  & 0.2984 & 57.23 & 43.45  & 0.1686 & - & - & -\tabularnewline
		T-GCN & 153.49 & 87.71  & 0.2918 & 56.35 & 43.28  & 0.1678 &129.56 & 85.66 & 0.2758\tabularnewline
		DRCNN & 151.59 & 86.14*  & 0.2884 & 54.73* & 41.81*  & 0.1637* &116.30 & 76.71&0.2561\tabularnewline
		ConSTGAT & 152.23 & 87.10  & 0.2845* & 56.67 & 43.22  & 0.1684 & 107.58* &69.90* &0.2288*\tabularnewline
		GCNAttTTE & 152.95 & 87.41  & 0.2898 & 55.88 & 42.79  & 0.1654 & 112.88 & 73.82 &0.2372\tabularnewline
		STDGNN & \textbf{135.50} & \textbf{74.40}  & \textbf{0.2524} & \textbf{50.51} & \textbf{38.35}  & \textbf{0.1537} & \textbf{90.49} & \textbf{55.24}  & \textbf{0.1920} \tabularnewline
		Improvements & \textbf{+8.63\%}  & \textbf{+13.88\%}  & \textbf{+11.28\%} & \textbf{+7.71\%}  & \textbf{+8.27\%}  & \textbf{+6.10\%} & \textbf{+15.88}\%& \textbf{+20.97\%}&\textbf{+16.08\%}\tabularnewline
		\bottomrule
  \end{tabular}}
  \vspace{-2mm}
\end{table*}

\begin{table*}[htb]
\centering
  \caption{Performance of STDGCN and other graph learning based methods for travel time estimation of the links on three datasets, and performance of STDGCN and GCNAttTTE for intersections travel time estimation on Beijing Dataset. Especially the predicting results for intersections are listed after the symbol '/'.} 
   \vspace{-3mm}
  \label{tab:overall_performance_link}
  \scalebox{0.9}{
  \begin{tabular}{c|ccc|ccc|ccc}
	\toprule
			  & \multicolumn{3}{c}{Chengdu} & \multicolumn{3}{c}{Porto} & \multicolumn{3}{c}{Beijing} \tabularnewline
			  Method & RMSE (sec) & MAE (sec) & MAPE & RMSE (sec)  & MAE (sec) & MAPE & RMSE (sec)  & MAE (sec) & MAPE\tabularnewline
		\midrule
		T-GCN & 57.84 & 24.85  & 0.5537 & 22.48 & 13.03  & 0.5122 &47.06/- &23.87/- &0.5886/-\tabularnewline
		DRCNN & 57.82 & 24.76  & 0.5513 & 20.59 & 11.68  & 0.4605 & 49.14/-& 26.23/- & 0.6366/-\tabularnewline
		ConSTGAT & 58.17 & 24.55  & 0.5503 & 21.76 & 12.32  & 0.4865 & 47.26/-& 21.64/-&0.5074/-\tabularnewline
		GCNAttTTE & 58.26 & 24.79  & 0.5512 & 19.89 & 10.95  & 0.4323 &45.31/22.88 & 21.50/10.12&0.4884/0.4665\tabularnewline
		STDGNN  & \textbf{55.16} & \textbf{22.45}  & \textbf{0.5063} & \textbf{15.01} & \textbf{7.46}  & \textbf{0.2786} & \textbf{35.81/18.64} & \textbf{12.84/5.88} & \textbf{0.2521/0.2410}\tabularnewline
		Improvements & \textbf{+5.40\%}  & \textbf{+8.15\%}  & \textbf{+3.61\%} & \textbf{+36.51\%}  & \textbf{+26.00\%}  & \textbf{+19.42\%} & \textbf{+33.01/10.24\%}& \textbf{+35.04/12.92\%}&\textbf{+26.46/17.51\%}\tabularnewline
		\bottomrule
  \end{tabular}}
  \vspace{-2mm}
\end{table*}

\subsection{Methods for Comparison}
To demonstrate the superiority of our model, we select ten state-of-art baselines to compare with STDGNN. These baselines can be divided into three following categories.

\noindent\textbf{Classical and statistics-based models:} 
\begin{itemize}
\item \textbf{AVG}: This is a classic method that has been deployed in multiple map services. The average speed of each link in the city during a specific time interval is calculated. The travel time of a given path can be estimated with the historical average speed and given departure time.
\item \textbf{TEMP~\cite{wang2019simple}}: This is a spatial aggregation method that adopts the historical trajectories with adjacent origin and destination to estimate travel time. In our experiment, the five most similar neighbor trajectories are assigned to each trajectory to be estimated.
\item \textbf{GBDT~\cite{friedman2001elements}}: This is an ensemble statistical learning method that uses gradient boosting decision regression tree for travel time estimation. Similar with~\cite{wang2018will}, we sample 128 GPS points for each sequence. And the max depth of the regression tree is set as 8.
\end{itemize} 

\noindent\textbf{Traditional deep learning based models:}
\begin{itemize}
\item \textbf{MlpTTE}: This is a simple deep learning model that adopts multiple-layer perceptron to estimate the travel time. In our experiments, the number of feed-forward fully connected layers is set as 3, and the activation function of each layer is selected as ReLU. The number of hidden units in MlpTTE is fixed as 64.
\item \textbf{RnnTTE}: Considering the temporal characteristic of the trajectory sequences, Gated Recurrent Unit Network (GRU) is adopted to process the raw GPS sequence into a 128-dimensional feature vector. Then, it is put forward into a fully connected layer to achieve the estimated results.
\item \textbf{DeepTTE~\cite{wang2018will}}: This is a hybrid end-to-end deep learning model that adopts Geo-Conv layer to capture spatial dependencies and GRU layer to capture the temporal dependencies. Similar with MlpTTE and RnnTTE, DeepTTE learns the spatio-temporal correlations based on the consecutive sampling GPS points along the query path. In the experiments, we set the hidden units of Geo-Conv layer and GRU layer as 64.
\end{itemize}

\noindent\textbf{Graph deep learning based models:}
\begin{itemize}
\item \textbf{T-GCN~\cite{zhao2019t}}: This is a hybrid GCN-based model that integrates GCN and GRU to simultaneously capture the spatial and temporal correlations in traffic flow prediction. In our experiments, its core architecture is introduced to learn the spatio-temporal representation for travel time estimation. The hidden size of GCN model and GRU are respectively set as 20 and 128.
\item \textbf{DCRNN~\cite{li2017diffusion}}: This is also a spatio-temporal graph-based deep framework in traffic prediction. This model exploits diffusion graph convolution to capture spatial dependencies, and then uses the recurrent neural networks to model temporal dependencies. Similar with T-GCN, we leverage its core architecture for travel time estimation in our experiments. The settings of the hidden size of GCN and GRU are the same as T-GCN model.
\item \textbf{ConSTGAT~\cite{fang2020constgat}}: This is a graph attention based model that adopts attention mechanism to extract the joint dependencies of spatial and temporal dynamics, which is superior to other traditional methods in travel time estimation. We implement its core architecture for spatio-temporal representation learning, where the size of the latent representation in this model is set as 32.
\item \textbf{GCNAttTTE}: This model takes both road segments and intersections into account. Different from STDGNN, we first use GCN for road segments modeling, and then involve attention mechanism to learn the latent features of intersections from the embeddings of connected road segments. In addition to the attention mechanism, the basic settings and the architectures of GCNAttTTE are the same as T-GCN model above. 
\end{itemize}

\begin{table*}[htb]
\centering
  \caption{Performance of STDGNN and baselines for global travel time estimation under three scenarios.} 
  \label{tab:overall_performance_case}
  \vspace{-3mm}
  \scalebox{0.95}{
  \begin{tabular}{c|ccc|ccc|ccc}
	\toprule
			  Scenario & \multicolumn{3}{c}{Suburb} & \multicolumn{3}{c}{Rush Hours} & \multicolumn{3}{c}{Non-Rush Hours}\tabularnewline
			  Method & RMSE (sec) & MAE (sec) & MAPE & RMSE (sec)  & MAE (sec) & MAPE & RMSE (sec)  & MAE (sec) & MAPE\tabularnewline
		\midrule
		AVG & 487.93 & 350.99  & 0.5372 & 338.77 & 251.29  & 0.7115 & 327.50 & 245.89  & 0.6985\tabularnewline
		T-GCN & 139.52 & 89.72  & 0.1454 & 139.21 & 85.12  & 0.2920 & 137.26 & 86.71  & 0.2867\tabularnewline
		DRCNN & 128.70 & 82.76  & 0.1318 & 132.36 & 83.43  & 0.2825 & 128.51 & 79.82  & 0.2701\tabularnewline
		ConSTGAT & 125.22 & 74.83  & 0.1165 & 129.88 & 80.04  & 0.2731 & 116.48 & 71.69  & 0.2584\tabularnewline
		GCNAttTTE & 131.01 & 77.25  & 0.1223 & 132.62 & 79.76  & 0.2716 & 120.41 & 74.35 & 0.2613\tabularnewline
		STDGCN (ours)  & \textbf{99.18} & \textbf{54.93}  & \textbf{0.0927} & \textbf{112.77} & \textbf{67.32}  & \textbf{0.2495}  & \textbf{98.62} & \textbf{64.25}  & \textbf{0.2359}\tabularnewline
		Improvements & \textbf{+20.80\%} & \textbf{+26.59\%}  & \textbf{+20.43\%} & \textbf{+13.17\%}  & \textbf{+15.60\%}  & \textbf{+8.14\%}  & \textbf{+15.33\%}  & \textbf{+10.37\%} & \textbf{+8.70\%} \tabularnewline
		\bottomrule
  \end{tabular}}
  \vspace{-3mm}
\end{table*}

\subsection{Experimental Results}
\subsubsection{Overall Performance} We compare our proposed model with other baselines on three datasets. Table~\ref{tab:overall_performance_path} shows the performance of the travel time estimation for the entire paths. Since the latitude and longitude of the Beijing dataset are desensitized, the models that use latitude and longitude as input cannot be used as comparative examples. 
Obviously, we find that the performance of our proposed model is superior to other methods in terms of all three metrics. 
To be specifical, our model outperforms at least 8.63\%, 13.88\%, 11.28\% in RMSE, MAE and MAPE on Chengdu dataset, and the improvements of these three metrics are at least 7.71\%, 8.27\%, 6.10\% on Porto dataset. On Beijing dataset, our model achieves more significant improvements, at least 15.88\%, 20.97\%, 16.08\% in RMSE, MAE and MAPE, which further demonstrates that STDGNN is an effective and practical solution for travel time estimation.
There are two main reasons to explain such improvements. First, some graph based methods like T-GCN, DCRNN and ConSTGAT only model the dynamics features of road segments. However, intersections are commonly the critical junctions to influence the traffic conditions among multiple connected links, which also play an significant role for estimating travel time. Therefore, designing a model that ignores intersections is not conducive to improving accuracy.
Second, for the models like GCNAttTTE, the features of intersections and links are both taken into account, but this model does not fully exploit the direct joint relations between the intersections and road segments along the path. Thus, GCNAttTTE also cannot obtain the optimal performance. 
Unlike these baselines, our proposed STDGNN not only considers both independent spatio-temporal correlations of intersections and links, but also captures their complex relations. Hence, it can outperform than other methods.

Because our model has the ability of multi-task learning, we conduct experiments to compare our model with other graph deep learning models on estimating travel time of the links and intersections. Note that, only Beijing Dataset provide the historical travel time of intersections, and no other baselines consider the travel time of intersections except for GCNAttTTE. Thus, we compare our model with GCNAttTTE for estimating travel time of the intersections only based on Beijing Dataset. 
As shown in Table~\ref{tab:overall_performance_link}, STDGNN obtains better performance of travel time estimation for links on all these three datasets. Also, there is a significant improvement by our model compared with GCNAttTTE, both for travel time of links and intersections.
This is because that our model only captures the independent spatio-temporal correlations of the links and intersections but also fully exploits the interaction relations between them, which simultaneously improve the accuracy of travel time estimation of links and intersections.

\subsubsection{Performance Improvement Under Different Scenarios} As we konw, both the spatial and temporal attributes have significant impacts on travel time estimation. To be specific, the traffic conditions vary with different spatial scopes. For instance, the busy traffic conditions are common in the downtown, while the traffic conditions are usually smooth in the suburb. In addition, traffic conditions are changing with time. For example, the probability of traffic congestion during the rush hours is much greater than during the non-rush hours. Based on these above common senses, we evaluate our model in travel time estimation under three different scenarios, including the suburb, rush hours and non-rush hours. 
Table~\ref{tab:overall_performance_case} shows the experiment results. As we know, during the rush hours, the occurrence of more traffic congestion cause difficulties in estimating travel time, our proposed model still improve at least 13.17\%, 15.60\% and 8.14\% in RMSE, MAE and MAPE. Also, our model obtains at least 15.33\%, 10.37\% and 8.70\% improvement in these three metrics for non-rush hours.
Meanwhile, in suburb with relatively sparse trajectories, our model achieves at least 20.80\%, 26.59\% and 20.43\% improvement of estimation in RMSE, MAE and MAPE. 
Obviously, regardless of which scenarios, our model significantly outperforms other methods. This demonstrates that the generalization of our model is satisfactory  for travel time estimation under different scenarios. 

\subsubsection{Ablation Study} 
To demonstrate that each module in our model is effective, we present six variants in ablation studies: 1) w/o Multi-scale , which removes the multi-scale architectures from spatio-temporal learning layer; 2) w/o TCN ,which removes TCN model from each spatio-temporal learning cell; 3) w/o GCN, which removes GCN model from each spatio-temporal learning cell; 4) w/o intersections, which only adopts the spatio-temporal features from the edge-wise graph; 5)w/o links, which only adopts the spatio-temporal features from the node-wise graph; 6) w/o P-matrix, which removes the incidence matrix $P$ in dual-graph interaction.

To validate how the multi-scale architectures in STDGNN can effectively capture the multi-scale spatio-temporal dynamics, we present the variant w/o Multi-scale. From Table~\ref{tab:ablation_study}, we find that the performance of this variants is worse than STDGNN on two datasets. For example, its performance degrades 5.41\%, 7.27\% and 5.98\% in terms of RMSE, MAE and MAPE on Chengdu dataset by the comparison with our model. This is because that the multi-scale architectures can better integrate the spatio-temporal dependencies from low-level to high-level. 
Then, we present w/o GCN and w/o TCN to validate the importance of the spatial and temporal learning part. The results are shown in Table~\ref{tab:ablation_study}, from which we find that the obvious declines occur in estimation performance on two datasets when removing the spatial or temporal learning part. This is because that both the spatial learning part and the temporal learning part play a central role in capturing the 
latent dynamics from road networks. Based on the above ablation variants, we conclude that each component in spatio-temporal learning cell and the multi-scale architectures can effectively capture the joint spatio-temporal correlations from the dual graphs.

\begin{table}[tb]
\centering
\caption{Performance of STDGNN and ablation variants for global travel time estimation on two datasets.} 
\label{tab:ablation_study}
\vspace{-3mm}
\scalebox{0.7}{
\begin{tabular}{c|ccc|ccc}
	\toprule
			  & \multicolumn{3}{c}{Chengdu} & \multicolumn{3}{c}{Porto}  \tabularnewline
			  Method & RMSE (sec) & MAE (sec) & MAPE & RMSE (sec)  & MAE (sec) & MAPE\tabularnewline
		\midrule
		w/o Multi-scale & 143.25 & 80.24  & 0.2685 & 51.43  & 40.75   & 0.1579\tabularnewline
        w/o GCN & 148.19 & 84.21  & 0.2776 & 53.78 & 42.03  & 0.1615\tabularnewline	
		w/o TCN & 146.76 & 83.37  & 0.2741 & 54.26 & 41.82  & 0.1623\tabularnewline
		w/o Intersection & 140.38 & 77.66  & 0.2626 & 52.57 & 40.83  & 0.1607\tabularnewline
		w/o Link & 144.56 & 81.39  & 0.2672 & 53.12 & 41.34  & 0.1625\tabularnewline
		w/o P-matrix & 138.95 & 76.58  & 0.2603 & 52.18 & 39.56  & 0.1572\tabularnewline
		STDGNN (ours)  & \textbf{140.30} & \textbf{77.60}  & \textbf{0.2585} & \textbf{50.51} & \textbf{38.35}  & \textbf{0.1537}\tabularnewline
		\bottomrule
  \end{tabular}}
  \vspace{-3mm}
\end{table}

Next, we investigate how intersection modeling, link modeling, and the joint interaction of the two affect estimation performance of the model. Hence, we respectively  present three variants, w/o intersections, w/o links and  w/o P-matrix.
From Table~\ref{tab:ablation_study}, we observe that the performance of these three variants become worse than our complete model regardless of which dataset. Specifically, the performance of w/o Intersections decreases 3.47\%, 4.19\% and 3.88\% on Chengdu dataset, and 3.91\%, 6.07\% and 4.35\% on Porto dataset, in terms of RMSE, MAE and MAPE. The performance of w/o Links decreases 6.26\%, 8.58\% and 5.53\% on Chengdu dataset, and 5.58\%, 7.34\% and 5.41\% on Porto dataset, in terms of RMSE, MAE and MAPE. The performance of w/o P-matrix decreases 2.48\%, 2.85\% and 3.04\% on Chengdu dataset, and 3.20\%, 3.06\% and 2.23\% on Porto dataset, in terms of RMSE, MAE and MAPE. This implies that both the independent characteristics of intersections and links or the interactive characteristics of the two have an crucial impact on travel time estimation. Thus, without any of them, our model cannot achieve optimal performance.

 \begin{figure}[tb]
 \centering
 \includegraphics[width=0.48 \textwidth]{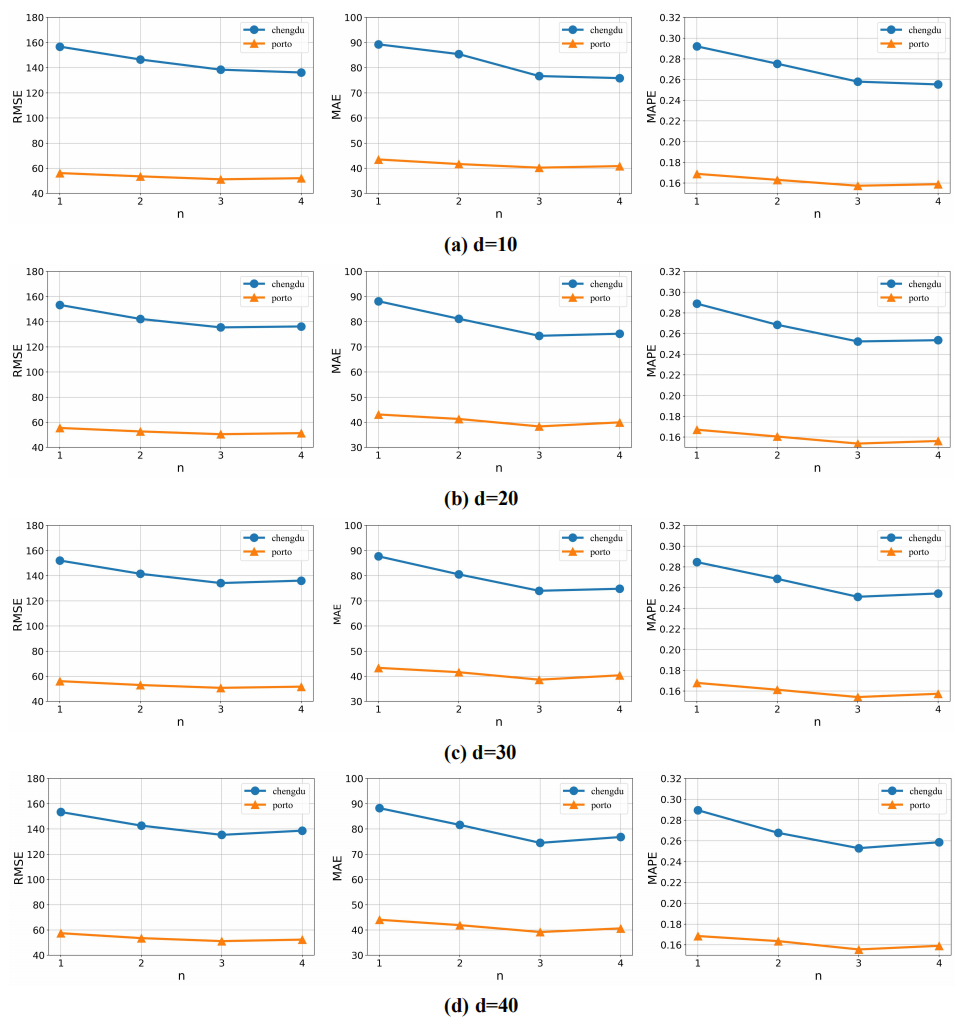}
 \caption{The trends of three different metrics of our model under different $n$. }
 \label{fig:parameter_analysis} 
 \end{figure}

 \begin{table}[tb]
\centering
  \caption{Performance of STDGNN under different combinations of $n$ and $d$.}
  \label{tab:parameter_analysis}
  \vspace{-3mm}
  \scalebox{0.8}{
  \begin{tabular}{c|ccc|ccc}
	\toprule
			   & \multicolumn{3}{c}{Chengdu} & \multicolumn{3}{c}{Porto} \tabularnewline
			  (n, d) & RMSE (sec) & MAE (sec) & MAPE & RMSE (sec)  & MAE (sec) & MAPE \tabularnewline
		\midrule
		(2, 10) &146.56	&85.42	&0.2753	&53.57	&41.68	&0.1631 \tabularnewline
		(2, 20) &142.24	&81.19	&0.2685	&52.71	&41.32	&0.1605 \tabularnewline
		(2, 30) &141.57	&80.55	&0.2684	&53.04	&41.62	&0.1613 \tabularnewline
		(2, 40) &142.68	&81.63	&0.2677	&53.57	&41.93	&0.1636 \tabularnewline
		(3, 10) &138.47	&76.69 &0.2580	&51.23	&40.24	&0.1574 \tabularnewline
		(3, 20) &135.50	&74.40	&0.2524	&\textbf{50.51}	&\textbf{38.35}	&\textbf{0.1537} \tabularnewline
		(3, 30) &\textbf{134.17}	&\textbf{74.01}	&\textbf{0.2511}	&50.78	&38.64	&0.1542  \tabularnewline
		(3, 40) &135.36	&74.51	&0.2530	&51.21	&39.20	&0.1556 \tabularnewline
		\bottomrule
  \end{tabular}}
  \vspace{-3mm}
\end{table}

\subsubsection{Parameter Sensitivity Study } 
To further investigate the parameter robustness  of our model, we design the comparative experiments under different combinations of some important parameters, including the number of layers of spatio-temporal cells (denoted by $n$) and the dimensions of latent representation (denoted by $d$).
Specifically, We divide the experiments into multiple groups based on fixed $d$ to evaluate our model under different $n$. As shown in Fig.~\ref{fig:parameter_analysis} (a)(b)(c)(d), we find that in terms of RMSE, MAE and MAPE, the performance gradually increases when $n$ goes from 1 to 3 under each $n$ on the two datasets. However, when $n=4$, the increment of the performance under $d=10$ slows down and the performance under other options of $d$ becomes worse. This reveals that more spatio-temporal cells does not necessarily mean the better performance. 
Further, to explore how the dimensions of latent representation affect the performance of our model, we fix $n$ and compare the performance under different $d$ on Chengdu and Porto dataset. The results with $n=[2,3]$ are shown in Table~\ref{tab:parameter_analysis}. We can find that no matter $k$ is 2 or 3, our model obtains the optimal performance on Chengdu dataset when $d=30$, while the optimal performance is obtained on Porto dataset when $d=20$. This indicates that larger $d$ may not bring better performance to the model. When $d$ increases moderately, the expressive capability of our model increases, but when $d$ become larger than a certain threshold, the problem of over-fitting could be caused to reduce the performance. 
To sum up, we conclude that the more spatio-temporal cells and the larger dimensions of latent representation does not mean the better performance. The specific value of them should be appropriately determined according to the corresponding dataset.

\section{Conclusion}
In this paper, we propose a novel spatio-temporal dual graph neural network framework named STDGNN for estimating travel time, which deals with the problem that road segments and intersections are not simultaneously taken into account by dual graph modeling.
To evaluate the effectiveness of our model, we conduct extensive experiments on three real-world datasets, and the experimental results show that STDGNN can achieve a higher accuracy in estimating the travel time of paths, as well as links and intersections. Meanwhile, the experiments also demonstrate the generality of STDGNN and the effectiveness of different components in this model.
However, there are still some limitations can be improved in the future. First, we do not involve some external data such as weather conditions and special events, which could have a significant impact on traffic states. Second, we only deploy our model on relatively small-scale road networks due to limited computing resources. In the future, we will consider more external factors that have a significant impact on travel time, and attempt to extend our model to larger-scale road networks.


\bibliographystyle{IEEEtran}
\end{document}